\documentclass[letterpaper, 10 pt, journal, twoside]{IEEEtran}
\usepackage{cite}

\ifCLASSINFOpdf
  \usepackage[pdftex]{graphicx}
\else
\fi
\usepackage{amsmath}
\usepackage{mathptmx}
\usepackage{amssymb}
\usepackage[cal=cm]{mathalfa}

\usepackage{array}

\usepackage{stfloats}
\usepackage{hyperref}
\usepackage{url}

\usepackage{makecell}
\usepackage{hhline}
\usepackage[table]{xcolor}
\usepackage{soul, color, xcolor}
\usepackage{multirow}

\begin{document}
%
\title{START: Traversing Sparse Footholds with \\Terrain Reconstruction}
%
%
%

\author{Ruiqi Yu$^{1}$, Qianshi Wang$^{1}$, Hongyi Li$^{2}$, Zheng Jun$^{3}$, Zhicheng Wang$^{1}$, Jun Wu$^{1}$, and Qiuguo Zhu$^{*1}$%
\thanks{Manuscript received: August, 18, 2025; Revised November, 14, 2025; Accepted December, 8, 2025.}
\thanks{This paper was recommended for publication by Editor Abhinav Valada upon evaluation of the Associate Editor and Reviewers' comments.
This work was supported by the "Leading Goose" R\&D Program of Zhe jiang (Grant No. 2023C01177), the National Key R\&D Program of China (Grant No. 2022YFB4701502), and the 2035 Key Technological Innovation Program of Ningbo City (Grant No. 2024Z300).}
\thanks{$^{1}$Authors are with Institute of Cyber-Systems and Control, Zhejiang University, 310027, China.
        {\tt\footnotesize \{yrq, niujiao, 3160105273\}@zju.edu.cn, jwu@iipc.zju.edu.cn, qgzhu@zju.edu.cn}.}%
\thanks{$^{2}$Author is with MARMoT Lab, National University of Singapore, 119077.
        {\tt\footnotesize hongyi.li@u.nus.edu}.}%
\thanks{$^{3}$Author is with the 19th Asian Games Hangzhou 2022 Organising Committee, 310020.
        {\tt\footnotesize zj558@126.com}.}%
\thanks{$^{*}$Qiuguo Zhu is the corresponding author.}%
\thanks{Supplementary video is available at \url{https://youtu.be/j85qVv_kVyI}.}%
\thanks{Digital Object Identifier (DOI): see top of this page.}
}
%
%

\markboth{IEEE Robotics and Automation Letters. Preprint Version. Accepted December, 2025}
{YU \MakeLowercase{\textit{et al.}}: START: Traversing Sparse Footholds with Terrain Reconstruction} 

%



\maketitle


\begin{abstract}
Traversing terrains with sparse footholds like legged animals presents a promising yet challenging task for quadruped robots, as it requires precise environmental perception and agile control to secure safe foot placement while maintaining dynamic stability. Model-based hierarchical controllers excel in laboratory settings, but suffer from limited generalization and overly conservative behaviors. End-to-end learning-based approaches unlock greater flexibility and adaptability, but existing state-of-the-art methods either rely on heightmaps that introduce noise and complex, costly pipelines, or implicitly infer terrain features from egocentric depth images, often missing accurate critical geometric cues and leading to inefficient learning and rigid gaits. To overcome these limitations, we propose START, a single-stage learning framework that enables agile, stable locomotion on highly sparse and randomized footholds. START leverages only low-cost onboard vision and proprioception to accurately reconstruct local terrain heightmap, providing an explicit intermediate representation to convey essential features relevant to sparse foothold regions. This supports comprehensive environmental understanding and precise terrain assessment, reducing exploration cost and accelerating skill acquisition. Experimental results demonstrate that START achieves zero-shot transfer across diverse real-world scenarios, showcasing superior adaptability, precise foothold placement, and robust locomotion.
\end{abstract}


%
\IEEEpeerreviewmaketitle

\section{Introduction}
%
%
%
%
\IEEEPARstart{L}{egged} animals demonstrate exceptional locomotion capabilities on diverse unstructured and complex terrains. Even in scenarios with extremely sparse footholds, such as stepping across streams on irregular stones or balancing on narrow branches, they traverse at high speed while maintaining stability. Unlike well-structured continuous terrains, where animals could rely solely on rapid proprioceptive feedback to perform dynamic locomotion, these discrete scenarios leave virtually no margin for error, even a slight misstep can lead to a fall \cite{farrell2015accurate}. Therefore, to achieve similar performance on risky terrains with sparse footholds, legged robots need the ability to leverage exteroception to accurately perceive their surroundings and employ tightly coupled feedforward control to adjust gait and body posture, ensuring each footfall lands within a safe region to maintain stable, agile locomotion \cite{loc2009control}.

\begin{figure}[!tbp]
  \centering
  \includegraphics[width=0.99\linewidth]{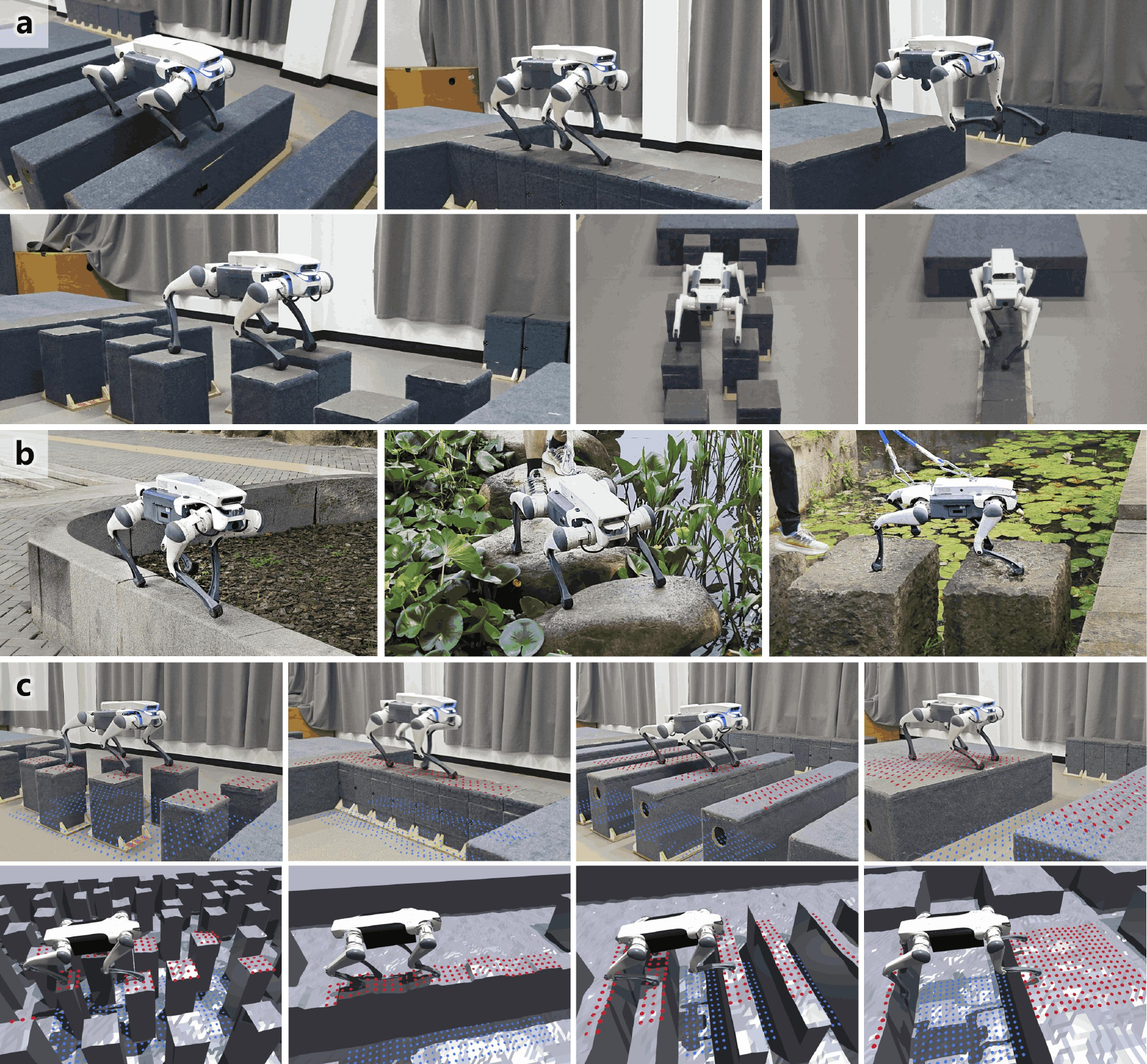}
  \vspace{-21pt}
  \caption{Our framework enables robot to perform agile and robust locomotion on challenging sparse foothold terrains, including stepping stones, balance beams, stepping beams, and gaps, with only onboard view-limited egocentric vision. The learned policy achieves zero-shot transfer to complex indoor and outdoor real world environments. (c) Precise local terrain reconstructions from TR-Net in the real world and simulation, accurately reproducing the geometric features essential for traversing sparse footholds.}
  \label{fig1}
  \vspace{-19pt}
\end{figure}

Model-based hierarchical controllers have been developed to tackle this challenge \cite{grandia2023perceptive,agrawal2022vision,griffin2019footstep,mastalli2020motion,fahmi2022vital}, decomposing sparse foothold locomotion tasks into three stages: perception, planning, and control \cite{yu2022visual}. Robots utilize simultaneous localization and mapping (SLAM) to generate heightmaps, solve motion trajectories via nonlinear optimization or graph search, and track them with low-level controllers \cite{margolis2022learning, yu2022visual,winkler2018gait,villarreal2020mpc,gangapurwala2022rloc}. Although effective in laboratory settings, these methods demand extensive hand-crafted design or parameter tuning, and the tradeoff between model fidelity and computational cost limits scalability to complex real-world scenarios \cite{margolis2022learning,yu2022visual,gangapurwala2022rloc}. Recent studies integrate learning-based planning and control methods to overcome these challenges \cite{yu2022visual,gangapurwala2022rloc,xie2022glide,tsounis2020deepgait}. For instance, Jenelten et al. \cite{jenelten2024dtc} combined model-based trajectory optimization for foothold planning with reinforcement-learning-based tracking, achieving state-of-the-art performance in challenging environments. Nevertheless, such decoupled planning-control architectures often require simplified, conservative assumptions to mitigate model mismatch, which could constrain each module's performance and yield suboptimal, overly conservative behaviors \cite{margolis2022learning,zhang2023learning,plagemann2008learning}.

To further unlock policy flexibility and adaptability, recent studies have developed model-free end-to-end reinforcement learning (RL) approaches to learn direct mappings from observations to joint actions, achieving significant progress in various structured environments \cite{miki2022learning, agarwal2023legged,cheng2024extreme,yang2021learning,yang2023neural,zhuang2023robot,hoeller2024anymal}. Nonetheless, locomotion over complex terrains with sparse, discrete footholds imposes stricter demands on timely, comprehensive, and precise terrain perception to inform action decisions. Since heightmaps theoretically offer a compact representation with clear physical meaning \cite{miki2022elevation}, Zhang et al. \cite{zhang2023learning} and He et al. \cite{he2025attention} have adopted local heightmaps as exteroception, enabling agile locomotion on risky sparse footholds. However, such clean heightmaps rely on external motion-capture (MoCap) systems for state estimation, restricting real-world applicability. To extend these methods outdoors, \cite{fankhauser2018probabilistic,fankhauser2016universal} perform model-based heightmap construction by fusing data from odometry and camera or LiDAR, but suffer from drift and estimation errors. Learning-based reconstruction methods address this by building heightmaps from noisy multi-view observations \cite{hoeller2022neural,hoeller2024anymal}, demonstrating superior results yet still requiring the robot’s estimated global pose to align consecutive frames. Moreover, these approaches often involve complicated hardware and introduce heavy computational load \cite{yang2023neural,agrawal2022vision}, unsuitable for low-cost robots.

In response to these limitations, researchers turn to egocentric vision, extracting environmental information directly from single-view depth images to infer and represent surrounding terrain. Most existing work, however, focuses on continuous terrains with height variations \cite{yang2021learning,loquercio2023learning}, such as stairs, slopes, or parkour obstacles, while research explicitly targeting extremely sparse footholds remains limited. We find that current approaches often use implicit vectors to encode terrain information \cite{agarwal2023legged,cheng2024extreme,yang2023neural,zhuang2023robot,luo2024pie}. Although this strategy increases information density and facilitates efficient signal propagation, the lack of explicit geometric supervision hampers the policy’s ability to accurately and comprehensively capture critical features necessary for traversing sparse footholds, impairing terrain evaluation. This imprecise feature representation often leads policies learning one single averaged gait pattern, restricting their adaptability to terrain sparsity and variability. Additionally, camera occlusions and limited viewpoints introduce partial observability, requiring the policy to maintain precise memory of past observations for reliable terrain assessment \cite{yang2023neural}.

In this paper, we propose \textbf{START}, a single-stage end-to-end learning framework that enables robust locomotion on terrains with sparse footholds such as stepping stones, balance beams, and stepping beams, using only view-limited egocentric vision, as shown in Fig. \ref{fig1}. To meet the strict demand for precise terrain perception, we introduce local terrain heightmap as an explicit intermediate representation between perception and output actions to encode critical geometric features (e.g., safe foothold areas and terrain edges) for stable locomotion. We develop a memory-augmented Terrain Reconstruction Network (TR-Net) that fuses proprioception with single-view depth images to reconstruct a local heightmap covering areas beneath and behind the robot, which serves as input to the subsequent locomotion policy. Experimental results indicate that, compared to implicit embeddings, our high-precision reconstruction captures and retains terrain features more comprehensively, enabling the policy to infer both its motion state and surroundings accurately. This facilitates safe and efficient locomotion on highly complex terrains with reduced exploration costs. Furthermore, local terrain reconstruction obviates the need for global mapping and localization to attain heightmaps, eliminating associated noise and drift. We also propose an Adaptive Sampling (AdaSmpl) method that tightly integrates the training of all network components into one single stage, significantly improving sample efficiency.

We deploy our policy on a low-cost DEEP Robotics Lite3 quadruped robot, and conduct extensive simulation and real-world experiments on challenging sparse footholds, demonstrating agile locomotion, precise foot placement, and robust stability. In summary, our main contributions are:

\begin{itemize}
\item \textbf{Efficient Learning for Risky Sparse Footholds:} We present a novel single-stage training framework that enables quadruped robot equipped solely with view-limited egocentric vision to learn safe locomotion over highly variable sparse footholds.
\item \textbf{Terrain Reconstruction:} We develop a memory-augmented TR-Net that generates precise local heightmaps to enhance environmental understanding and accurate foot placement, combined with AdaSmpl to boost sample efficiency and policy performance.
\item \textbf{Real World Generalization:} Our learned policy demonstrates high-agility locomotion in the real world and achieves zero-shot transfer to diverse outdoor environments, validating its robustness and generalizability.
\end{itemize}

%

\vspace{-5pt}
\section{METHOD}

\begin{figure*}[!htbp]
  \centering
  \includegraphics[width=0.84\linewidth]{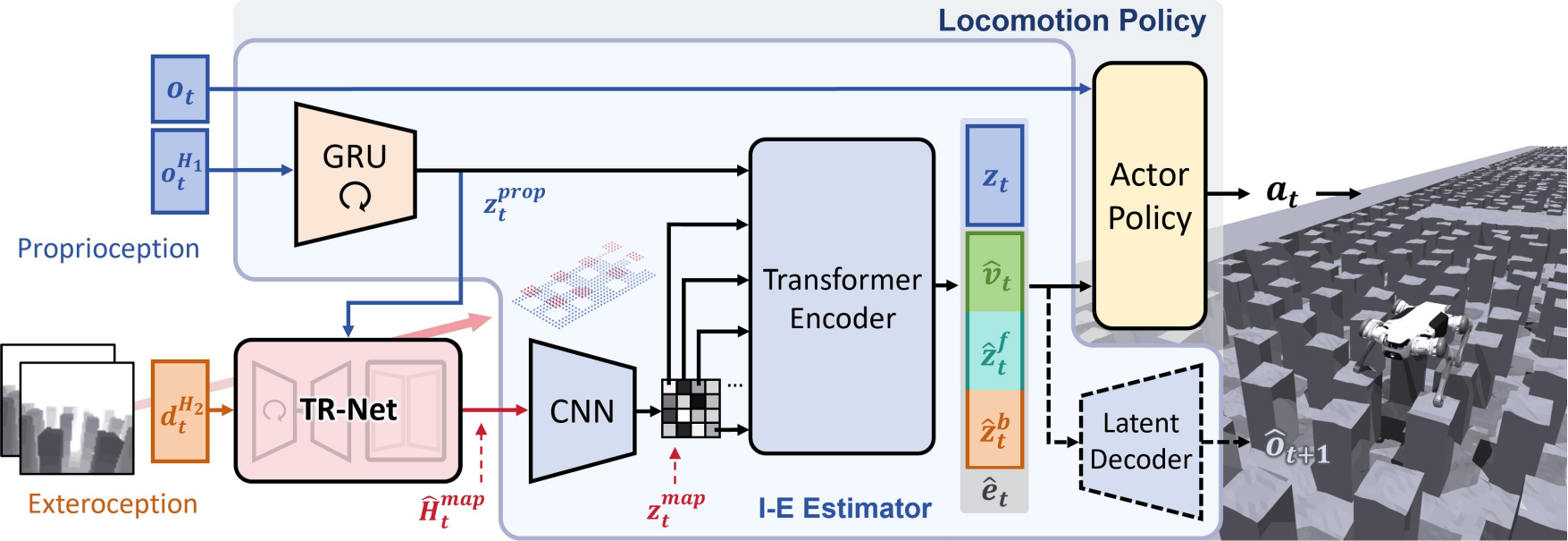}
  \vspace{-10pt}
  \caption{Overview of the START framework. The single-stage training framework is composed of: a \textcolor[rgb]{0.784, 0.114, 0.192}{\textbf{TR-Net}} that receives proprioceptive features and depth images to reconstruct local heightmap in real time, conveying precise terrain features of sparse footholds to the Locomotion Policy; a \textcolor[rgb]{0.1255, 0.2235, 0.4235}{\textbf{Locomotion Policy}} leverages an Implicit-Explicit Estimator (\textcolor[rgb]{0.192, 0.341, 0.635}{\textbf{I-E Estimator}}) to further enhance environmental understanding and generate actions with safe foot placements.}
  \label{fig2}
  \vspace{-20pt}
\end{figure*}

Our end-to-end learning-based framework START utilizes a set of neural networks to generate joint targets directly from proprioception and raw depth images, with local heightmap serving as an explicit intermediate representation. It comprises two core components that are jointly optimized in a single-stage pipeline, as shown in Fig. \ref{fig2}. The locomotion policy takes the reconstructed local heightmap and proprioception to generate joint actions (Section \ref{sec2-a}). Upstream, TR-Net reconstructs an accurate local heightmap for the policy from view-limited vision and proprioceptive features, clearly encoding the critical geometric cues of sparse footholds with high fidelity (Section \ref{sec2-b}). We also integrate the AdaSmpl method to enhance learning efficiency (Section \ref{sec2-c}).

\vspace{-6pt}
\subsection{Terrain-Aware Locomotion Policy}
\label{sec2-a}

The locomotion policy consists of an Implicit-Explicit Estimator (I-E Estimator) and an Actor Policy. The I-E Estimator, trained via supervised learning, produces both implicit and explicit estimation vectors that feed into the actor policy for action generation. We optimize the actor policy using proximal policy optimization (PPO) \cite{schulman2017proximal}. To avoid potential information gap introduced by the two-stage training paradigm \cite{fu2023deep,nahrendra2023dreamwaq,ji2022concurrent}, we adopt an asymmetric actor-critic architecture following \cite{nahrendra2023dreamwaq} for single-stage training.

\subsubsection{Observations and Action}

The actor policy consumes the latest proprioception $o_t$, the explicit estimation $\hat e_t$, and the implicit encoding $z_t$ from I-E Estimator, and outputs joint position targets $a_t \in \mathbb{R}^{12}$ for the 12 joints of the quadruped robot. The proprioception $o_t \in \mathbb{R}^{45}$ is measured directly from joint encoders and IMU, defined as:
$$
\setlength{\abovedisplayskip}{0pt}
\setlength{\belowdisplayskip}{0pt}
o_t=[\omega_t\quad g_t\quad cmd_t\quad \theta_t\quad \dot\theta_t\quad a_{t-1}]^T, \eqno{(1)}
$$
where $\omega_t \in \mathbb{R}^{3}$, $g_t  \in \mathbb{R}^{3}$, $cmd_t  \in \mathbb{R}^{3}$, $\theta_t  \in \mathbb{R}^{12}$, $\dot\theta_t  \in \mathbb{R}^{12}$, and $a_{t-1}  \in \mathbb{R}^{12}$ denote body angular velocity, gravity vector in the robot frame, velocity command, joint angle, joint angular velocity, and the previous action, respectively. The critic policy receives not only $o_t$ but also privileged environmental information $s_t$, which is defined as:
$$
\setlength{\abovedisplayskip}{0pt}
\setlength{\belowdisplayskip}{0pt}
s_t=[v_t\quad c_t\quad H^b_t\quad H^f_t]^T, \eqno{(2)}
$$
where $v_t$, $c_t$, $H^b_t$, and $H^f_t$ denote the ground-truth base velocity, foot-ground contact state, and heightmaps around the robot’s body and four feet, respectively.

\subsubsection{I-E Estimator} 

Robots can infer their own state and surroundings from observations \cite{kumar2021rma}. By guiding both implicit and explicit estimation, they can better extract privileged information to enhance environmental awareness and adaptability \cite{nahrendra2023dreamwaq,wang2024toward}. Based on this insight, we adopt a multi-head autoencoder–based I-E Estimator that fuses proprioception and exteroception to estimate robot state and privileged terrain information. Within I-E Estimator, a gate recurrent unit (GRU) processes temporal proprioception observations $o^{H_1}_t$ to produce proprioceptive features $z^{prop}_t$, while a convolutional neural network (CNN) encoder extracts terrain features $z^{map}_t$ from reconstructed local heightmap $\hat H^{map}_t$. A set of transformer encoder layers then fuse these features into cross-modal estimation vectors, implicit latent $z_t$ and explicit encoding $\hat e_t$, for the actor policy. Since each terrain token corresponds to a specific local region, self-attention enables spatial reasoning and focus on relevant areas \cite{yang2021learning}.

The explicit encoding is defined as $\hat e_t=[\hat v_t\enspace \hat z^b_t\enspace \hat z^f_t]^T$, where the base velocity $\hat v_t$ provides the robot’s motion state, $\hat z^b_t$ and $\hat z^f_t$ represents the explicit descriptors for body-centric and foot-centric heightmaps to enhance terrain awareness, and will be further decoded into $\hat H^b_t$ and $\hat H^f_t$ (within 0.1m), respectively. In contrast to \cite{luo2024pie}, which predicts only foot clearance, our per-foot heightmap estimation captures each footstep's positions and detailed terrain context for edge-avoiding foot placements. During training, all estimations are supervised against privileged ground-truth with a hybrid loss:
\begin{equation}
\setlength{\abovedisplayskip}{1pt}
\setlength{\belowdisplayskip}{1pt}
\begin{aligned}
\mathcal{L}_{\text{IE}} &= D_{\text{KL}}(q(z_t|o^{H_1}_t,\hat H^{map}_t)||p(z_t))+\text{MSE}(\hat o_{t+1}, o_{t+1})\\
&\quad +\text{MSE}(\hat v_t, v_t)+\text{MSE}(\hat H^f_t, H^f_t)+\text{MSE}(\hat H^b_t, H^b_t),
\end{aligned}
\tag{3}
\end{equation}
where the implicit encoding $z_t$ is obtained by a variational autoencoder (VAE) structure, and $\hat o_{t+1}$ is predicted from all estimation vectors to match the next-step proprioception $o_{t+1}$.

\begin{figure}[!tbp]
  \centering
  \vspace{-1pt}
  \includegraphics[width=0.98\linewidth]{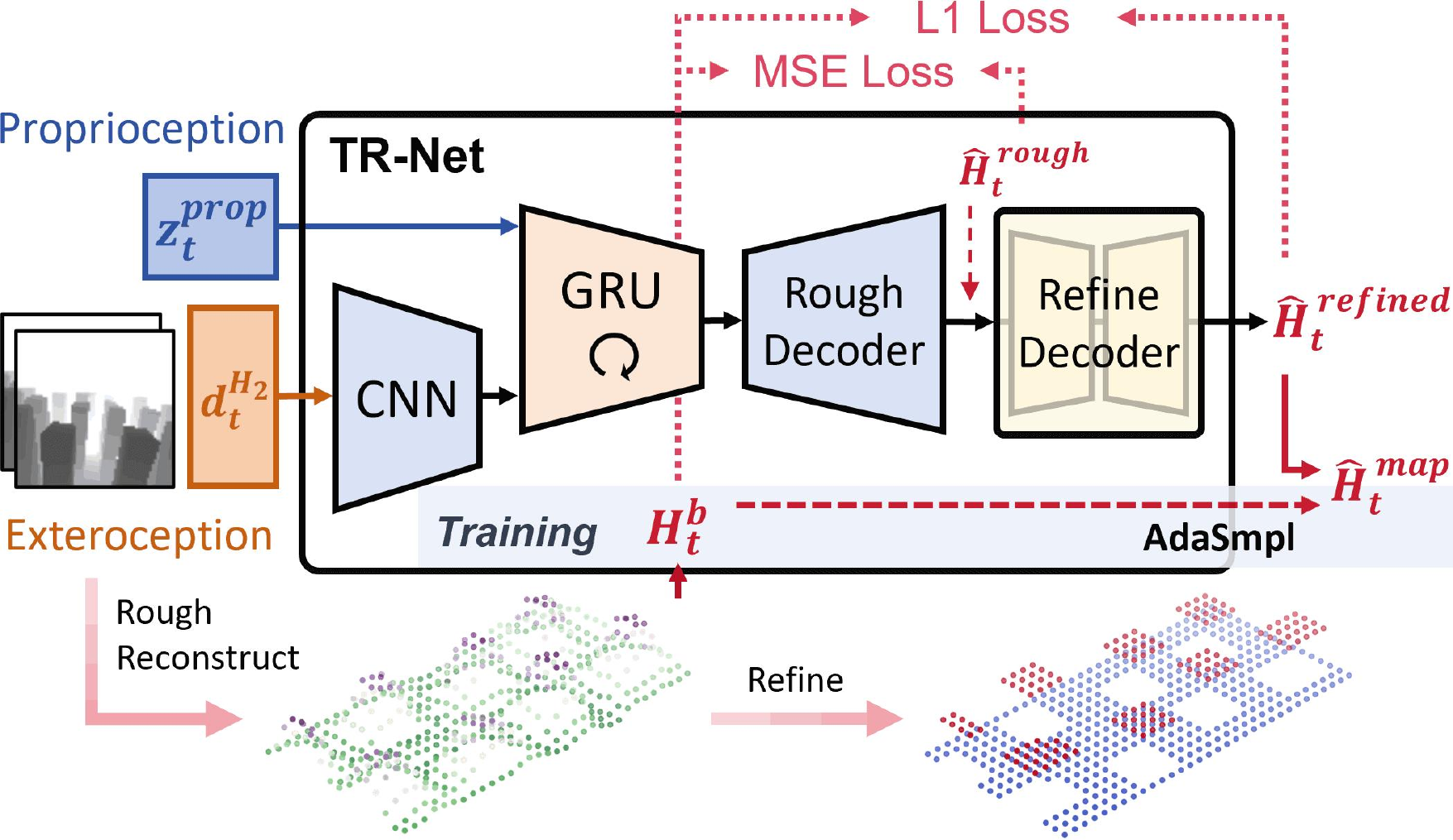}
  \vspace{-11pt}
  \caption{TR-Net extracts and retains essential visual information, fusing them with proprioceptive features to reconstruct and refine the local terrain heightmap that includes regions beneath and behind the robot. During training, we employ AdaSmpl method to enhance the exploration efficiency of the locomotion policy.}
  \label{fig3}
  \vspace{-18pt}
\end{figure}

\vspace{-8pt}
\subsection{TR-Net} 
\label{sec2-b}

TR-Net receives proprioceptive feature $z^{prop}_t$ and temporal depth images $d^{H_2}_t$ ($H_2=2$) as inputs, and reconstructs a precise heightmap in the robot’s local frame. The heightmap spans from 0.5m behind to 1.1m in front of the robot, with a width of 0.8m and a spatial resolution of 5cm. Examples of reconstruction results in the real world and simulation are illustrated in Fig. \ref{fig1}(c).

As shown in Fig. \ref{fig3}, TR-Net first applies a CNN encoder to extract visual features from depth images $d^{H_2}_t$, which are concatenated with proprioceptive features $z^{prop}_t$ that encodes the robot’s motion state. The combined vector is then fed to a GRU module to retain information about terrain currently occluded beneath and behind the robot. A multi-layer perceptron (MLP) decoder subsequently maps the integrated representation to a rough heightmap $\hat H^{rough}_t$. To further improve reconstruction accuracy, inspired by Duan et al. \cite{duan2024learning}, we incorporate a U-Net–based refinement decoder to recover high-frequency geometric details while suppressing noise and irregular, blurry edges, producing a refined heightmap $\hat{H}^{refined}_t$. We supervise $\hat H^{rough}_t$ with mean squared error (MSE) loss against the ground-truth $H^{b}_t$, while the refined reconstruction $\hat{H}^{refined}_t$ is optimized with an L1 loss to mitigate the over-smoothing tendency of MSE and to improve accuracy at edges and flat surfaces, thereby reducing the burden on the locomotion policy for terrain feature extraction. The loss for TR-Net is defined as:
$$
\setlength{\abovedisplayskip}{0pt}
\setlength{\belowdisplayskip}{0pt}
\mathcal{L}_{\text{TR}}=\text{MSE}(\hat H^{rough}_t, H^b_t)+\text{L1}(\hat H^{refined}_t, H^b_t). \eqno{(4)}
$$

\begin{table}[!tbp]
\setlength{\abovecaptionskip}{0pt}
\setlength{\belowcaptionskip}{0pt}
\caption{Detail reward terms.}
\vspace{-8pt}
\label{table1}
\begin{center}
\renewcommand\arraystretch{1.14}
\begin{tabular}{lll}
\Xhline{0.75pt}
{Reward Term}      & Equation                                 & Weight              \\
\hline
Lin. vel. tracking & $\text{exp}(-||\text{min}(v, v^{\text{cmd}})-v^{\text{cmd}}||^2/0.25)$         & 1.5 \\
\noalign{\vskip -1pt}
Ang. vel. tracking & $\text{exp}(-||\omega_{yaw}-\omega^{\text{cmd}}_{\text{yaw}}||^2/0.25)$ & 0.5       \\
\noalign{\vskip -1pt}
Lin. vel. (z)      & $v_z^2$                                  & -2.0                \\
\noalign{\vskip -1pt}
Ang. vel. (xy)     & $||w_{xy}||^2$                           & -0.05               \\
\noalign{\vskip -1pt}
Orientation        & $g_x^2+g_y^2$                            & -1.0                \\
\noalign{\vskip -1pt}
Torques            & $\sum_{j\in \text{joints}}|\tau_j|^2$           & -1.0$\times10^{-5}$ \\
\noalign{\vskip -1pt}
Action rate        & $\sum_{j\in \text{joints}}|a_t-a_{t-1}|^2$                & -0.01               \\
\noalign{\vskip -1pt}
Smoothness         & $\sum_{j\in \text{joints}}|a_t-2a_{t-1} + a_{t-2}|^2$     & -0.01               \\
\noalign{\vskip -1pt}
Joint power        & $\sum_{j\in \text{joints}}|\tau_j||\dot{q_j}|$  & -2.0$\times10^{-5}$ \\
\noalign{\vskip -1pt}
Joint accelerations & $\sum_{j\in \text{joints}}|\ddot{q_j}|^2$       & -2.5$\times10^{-7}$ \\
\noalign{\vskip -1pt}
Joint error        & $\sum_{j\in \text{joints}}|q_j-q^{\text{default}}_j|^2$          & -0.01               \\
\noalign{\vskip -1pt}
Collision          & $\sum_{i\in \text{contact}}\mathbf{1}\{F_i>0.1\}$         & -10.0               \\
\noalign{\vskip -0.5pt}
Stumble            & $\mathbf{1}\{\exists i,|F^{xy}_i|>4|F^z_i|\}$      & -1.0                \\
\noalign{\vskip -1pt}
Feet edge          & $\sum_{i\in \text{feet}}c_i\cdot\sum_{d\in \text{dist}} \omega_d E_d[p_i]$     & -1.0                \\ 
\Xhline{0.75pt}
\end{tabular}
\end{center}
\vspace{-25pt}
\end{table}

\vspace{-10pt}
\subsection{Adaptive Sampling}
\label{sec2-c}

Although TR-Net enhances accurate feature representation, its early outputs often remain noisy and incomplete, with invalid data obscuring clear geometric cues. To reduce early-stage exploration cost of the locomotion policy and improve sample efficiency, we introduce Adaptive Sampling (AdaSmpl), which probabilistically substitutes reconstructed heightmaps with ground-truth as inputs to the locomotion policy during training. The sampling probability is adapted based on training performance:
$$
\setlength{\abovedisplayskip}{0pt}
\setlength{\belowdisplayskip}{0pt}
p_{\text{smpl}}=\text{tanh}(CV(R)), \eqno{(5)}
$$
where $CV$ denotes the coefficient of variation of the episode rewards $R$. High $CV$ indicates unstable performance and triggers more frequent ground-truth sampling, providing privileged terrain information to ease sparse-reward exploration in complex environments. As training stabilizes and $CV$ decreases, the policy gradually relies more on reconstructed heightmaps. This schedule progressively exposes the policy to realistic, unstructured reconstruction noise, thereby enhancing robustness against noise and preventing performance degradation upon sim-to-real transfer. During deployment, the locomotion policy uses only reconstructed heightmaps.

\begin{figure}[!tbp]
  \centering
  \includegraphics[width=0.93\linewidth]{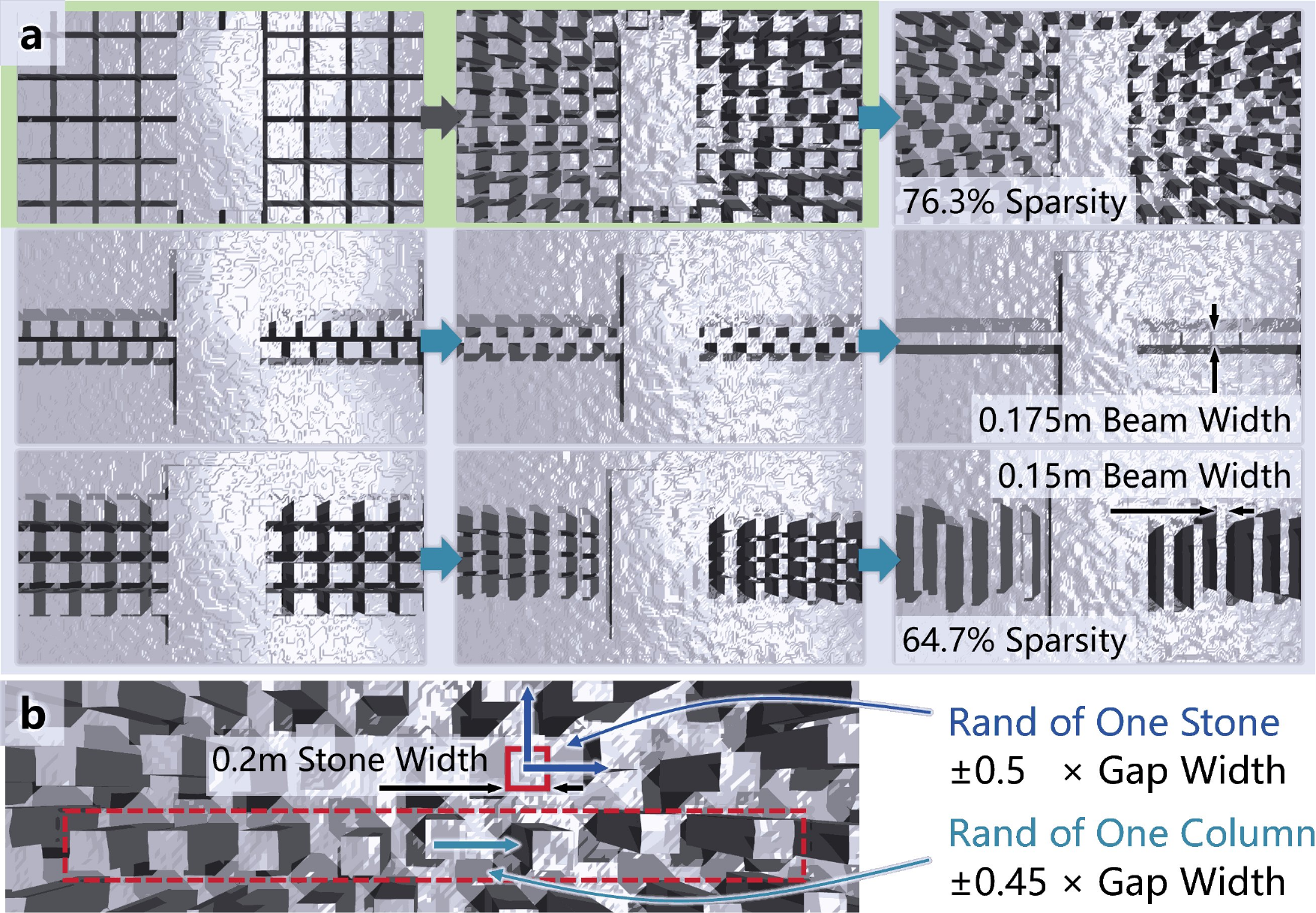}
  \vspace{-10pt}
  \caption{(a) Robots initially train on lower-randomness stepping stones (light green) to acquire basic motor skills and then progressively transition to more challenging sparse foothold terrains (light blue). (b) For higher-randomness stepping stones, we implement two levels of stone position randomization.}
  \label{fig4}
  \vspace{-17pt}
\end{figure}

\vspace{-8pt}
\subsection{Training Environment}
\label{sec2-d}

\subsubsection{Training Details}

We conduct single-stage training on Isaac Gym \cite{makoviychuk2021isaac} using 3072 parallel instances of DEEP Robotics Lite3 quadruped robot on one NVIDIA RTX A6000 GPU for approximately 10000 iterations/ 16.6 hours. To accelerate training, depth images at $60\times 60$ resolution are rendered with NVIDIA Warp instead of the default camera interface \cite{macklin2022warp}. We reset any robot that flips beyond predefined angular limits or whose foot height drops below a specified threshold beneath the ground plane. We also apply domain randomization following \cite{luo2024pie} to improve policy robustness.

Since terrains align along the x axis, we sample forward velocity commands uniformly from $[0.0, 1.5]\text{m/s}$ in the robot frame, with the lateral velocity fixed at zero. To enable turning, we clamp forward commands below 0.3m/s to zero, then sample an angular velocity from $[-1.2, 1.2]\text{rad/s}$ when the command is zero. This configuration trains a policy capable of simultaneous forward walking and turning.

\subsubsection{Rewards}

We adopt relatively simple reward functions following \cite{rudin2022learning} to demonstrate our framework enhances policy performance on sparse footholds. To encourage foot placements within safe regions, we introduce an edge-penalty:
$$
\setlength{\abovedisplayskip}{1pt}
\setlength{\belowdisplayskip}{1pt}
r_{\text{feetedge}}=-\sum_{i\in \text{feet}}c_i\cdot\sum_{d\in \text{dist}} \omega_d \cdot E_d[p_i], \eqno{(6)}
$$
where $c_i$ is the contact state, and $E_d$ is a boolean mask indicating whether foot position $p_i$ lies within $d$cm from an edge. We set $\text{dist}=[2.5, 5.0]$cm with penalty weights $\omega_{d}=[1.0, 0.5]$. The complete rewards are listed in Table \ref{table1}.

\subsubsection{Terrains and Terrain Progressive Curriculum}

We generate four types of terrain during training, each with 10 difficulty levels, and increase difficulty following the curriculum in \cite{rudin2022learning}. These terrains include three archetypal sparse foothold scenarios: stepping stones, balance beams, and stepping beams, as shown in Fig. \ref{fig4}. The stepping stones feature up to 76.3\% sparsity\footnote{Sparsity is defined as the ratio of the non-steppable area to the total terrain area\cite{zhang2023learning}.}, with stones 0.2m wide and gaps of 0.25m along x and 0.175m along y. Balance beams are as narrow as 0.175m. Stepping beams reach up to 64.7\% sparsity, with 0.15m beam widths and 0.275m gaps. We also include gaps up to 0.7m wide ($2\times$ robot length). Each patch measures $8\text{m}\times 4\text{m}$. To promote adaptability, we sample terrain depths from $[0.2, 0.7]$m and apply height variations up to ±5cm to all stones and beams.

To further reduce the difficulty of early-stage exploration and accelerate adaptive behaviors learning, we introduce an automatic terrain progressive curriculum, inspired by Zhang et al. \cite{zhang2023learning}. We categorize stepping stones into low- and high-randomness groups. Next, we incorporate balance beams and stepping beams as progressive transitions beyond stepping stones, as shown in Fig. \ref{fig4}(a). Initially, robots on sparse foothold terrains train on low-randomness stepping stones and auxiliary flat ground to acquire basic locomotion skills. As training advances, they are probabilistically allocated to high-randomness stepping stones, balance beams, and stepping beams. The probability follows a linear schedule: $p_{\text{advance}}=\text{max}(\text{min}(\frac{p_{\text{max}}(T-T_{\text{start}})}{T_{\text{end}}-T_{\text{start}}}, 0), p_{\text{max}})$. Gaps are excluded from this progression. Ultimately, the policy learns adaptive locomotion skills across all challenging terrains.

To ensure robustness on stepping stones with high sparsity and randomness, we randomize stone positions by shifting each stone along both axes by up to $0.5\times$ the gap and offsetting entire columns along the x-axis by up to $0.45\times$ the gap, as shown in Fig. \ref{fig4}(b).

\vspace{-6pt}
\section{EXPERIMENT}

\vspace{-3pt}
\subsection{Ablation Studies}
\vspace{-3pt}

We compare our START framework against eight ablation methods. The first group evaluates model architecture:

\begin{itemize}
\item \textbf{HMap GT}: This policy feeds the locomotion policy with ground-truth heightmap as input, with all other settings identical to START, representing an upper-bound performance. It provides a comparison using identical input sources to methods that obtain heightmaps via mapping or external MoCap \cite{zhang2023learning, he2025attention}.

\item \textbf{PIE (START w/o TR-Net)}: This policy removes TR-Net and encodes terrain information implicitly with a 2D-GRU. Its network structure is identical to PIE \cite{luo2024pie}, providing a fair comparison. It also assesses the efficacy of explicit heightmap reconstruction for conveying key terrain geometry.


\item \textbf{TR-Net w/o Prop}: This policy omits proprioceptive feature input in TR-Net and relies only on depth images.

\item \textbf{TR-Net w/o GRU}: This policy replaces the GRU in TR-Net with an MLP encoder to disable temporal memory.

\item \textbf{TR-Net w/o Refine}: This policy skips U-Net–based refinement in TR-Net and feeds the rough heightmap directly to the locomotion policy.
\end{itemize}

The second group evaluates training pipeline components:

\begin{itemize}
\item \textbf{START w/o AdaSmpl}: This policy is trained without the AdaSmpl method, so the locomotion policy always uses reconstructed heightmaps as input.

\item \textbf{START w/o FtEdge}: This policy is trained without foot-edge penalty term $r_{\text{feetedge}}$.

\item \textbf{START w/o TerProg}: This policy is trained without the terrain progressive curriculum and is exposed to all highly risky sparse foothold terrains from scratch.
\end{itemize}

We exclude implicit-explicit estimation from ablation study as it was already evaluated in prior work \cite{wang2024toward,luo2024pie}. We use three metrics: (1) \textbf{success rate}: the proportion of runs that successfully complete the full distance, evaluated per difficulty level; (2) \textbf{traversing rate}: the ratio of distance covered before falling to total distance, evaluated per difficulty level; and (3) \textbf{mean edge violation} (\textbf{MEV}): the proportion of steps landing on terrain edges, measured at the highest difficulty level. For methods involving TR-Net, we also measure reconstruction errors in simulation. All experiments were conducted with 500 runs over a 6m traversal on 10 randomly generated terrains.

\vspace{-8pt}
\subsection{Simulation Experiments}
\vspace{-3pt}

\subsubsection{Terrain Reconstruction Accuracy}

\begin{figure}[!tbp]
  \centering
  \includegraphics[width=0.93\linewidth]{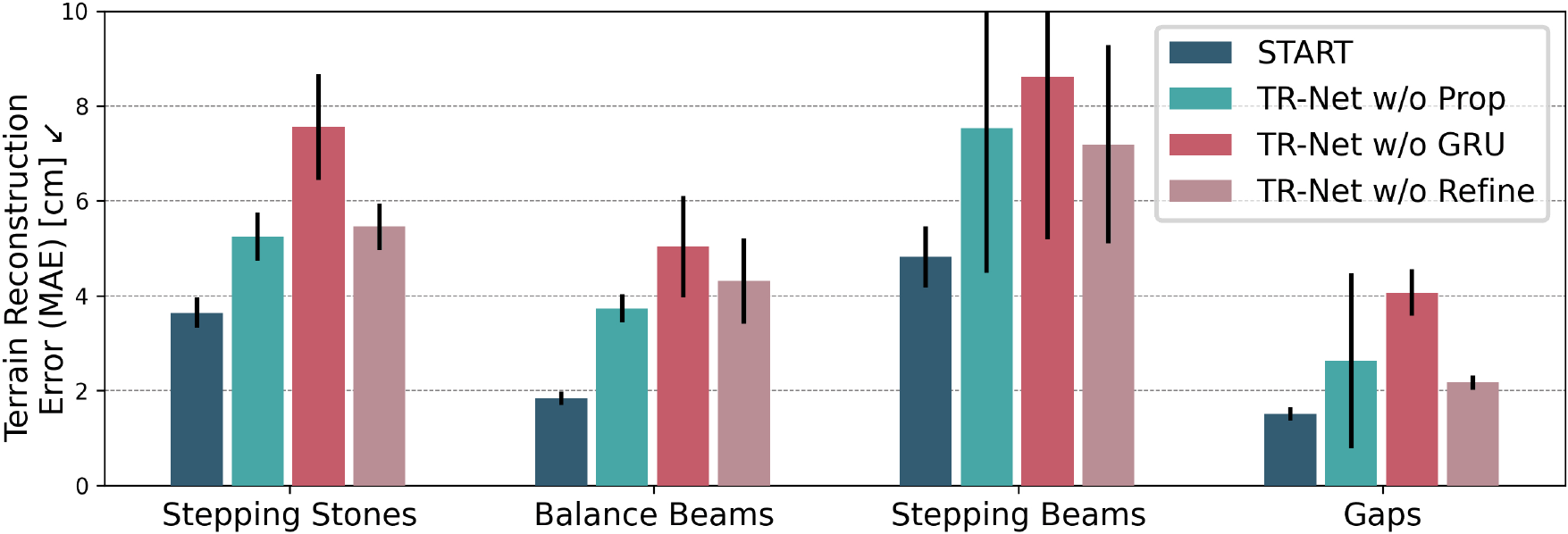}
  \vspace{-10pt}
  \caption{Terrain reconstruction error for various TR-Nets at the highest terrain difficulty level. Higher errors indicate poorer reconstruction performance.}
  \label{fig5}
  \vspace{-18pt}
\end{figure}

We assess the reconstruction performance of TR-Net variants. As shown in Fig. \ref{fig5}, lacking temporal memory, TR-Net w/o GRU exhibits the highest error, since it can only reconstruct terrain visible from the current viewpoint. TR-Net w/o Prop, which receives only depth images, performs only slightly better than the former. It neglects important motion states (e.g., base velocity) from proprioceptive features, impairing its ability to correlate current observation with past reconstructions and update its memory accurately. TR-Net w/o Refine omits refinement process and relies solely on a rough decoder, yielding noisy outputs with irregular edges. This limitation stems from restricted spatial awareness and poor high-frequency detail modeling of a single MLP, preventing accurate edge geometry capture. In contrast, our full TR-Net fuses proprioception and vision, leverages GRU-based temporal context and U-Net spatial-detail refinement, thereby achieving superior reconstruction accuracy across all complex terrains.

Notably, the focus of terrain reconstruction in our work is to help enhance robot locomotion, rather than mapping itself. By framing terrain reconstruction as an explicit terrain-feature extraction task, we produce an intermediate representation that connects accurate perception from limited egocentric vision with rapid, safe actions. Compared to approaches like \cite{hoeller2022neural,fankhauser2018probabilistic,fankhauser2016universal}, START requires only single-view depth images from a low-cost onboard camera, thereby avoiding the complexities of mapping and global estimation.

\begin{figure*}[!htbp]
  \centering
  \includegraphics[width=0.92\linewidth]{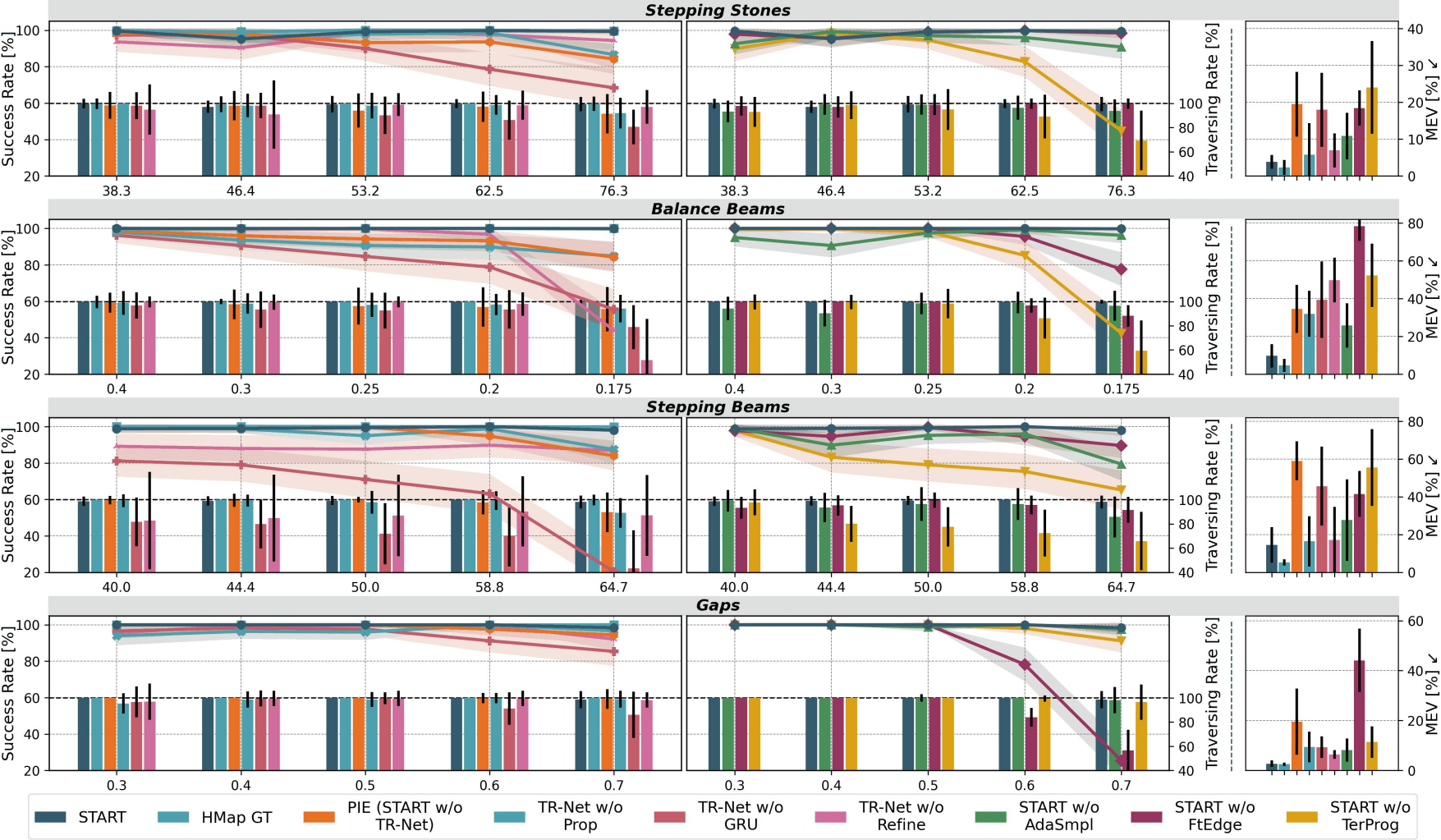}
  \vspace{-10.5pt}
  \caption{Ablation studies in simulation. The \textbf{left} and \textbf{middle} columns show \textbf{success rate} (line) and \textbf{traversing rate} (bar) for model-architecture and training-pipeline groups, both plotted against terrain difficulty. Difficulty is defined by sparsity (with maximum randomization) for stepping stones and stepping beams, and by beam and gap width for balance beams and gaps. The \textbf{right} column reports \textbf{mean edge violations} (\textbf{MEV}) at the highest difficulty level.}
  \label{fig6}
  \vspace{-18pt}
\end{figure*}

\begin{figure}[!tbp]
  \centering
  \vspace{-2pt}
  \includegraphics[width=0.84\linewidth]{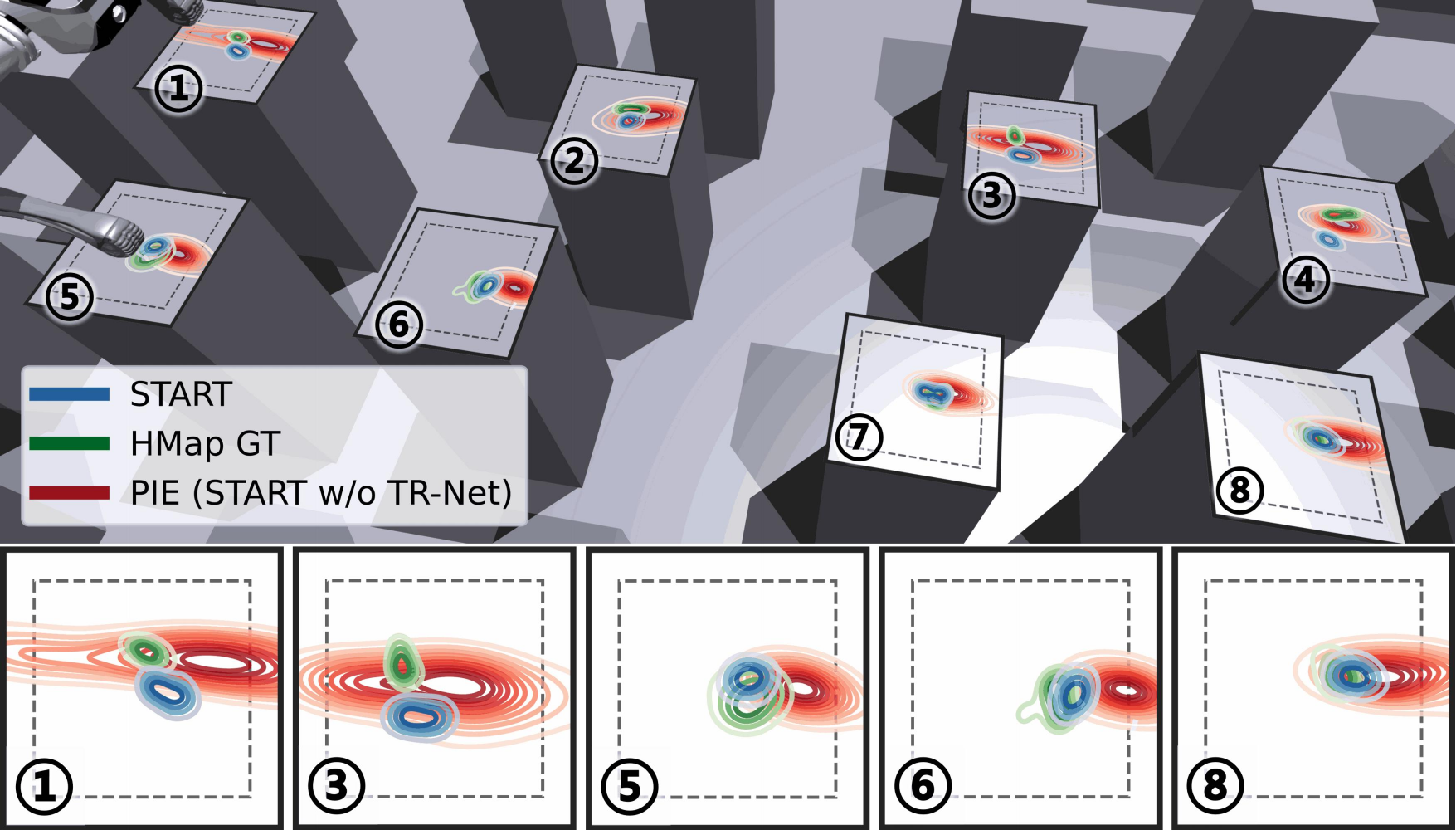}
  \vspace{-10pt}
  \caption{Foothold locations distribution for START, HMap GT, and PIE (START w/o TR-Net) on the highest difficulty level of stepping stones. We conducted 500 runs per policy and converted recorded footholds into continuous distributions via kernel density estimation (KDE).}
  \label{fig7}
  \vspace{-21pt}
\end{figure}

\subsubsection{Policy Performance}

Results in Fig. \ref{fig6} demonstrate the advantages of our START framework over all ablation methods across diverse terrains. First, it underscores the importance of explicit local heightmaps reconstructed by TR-Net for accurately encoding foothold-relevant terrain features, which enable stable traversal on sparse footholds. In contrast, PIE (START w/o TR-Net) relies on implicit feature vectors extracted directly from depth images. These vectors lack the clear physical meaning of heightmaps and thus struggle to capture precise geometric details. Consequently, the robot’s perception of rapidly changing terrain is impaired, causing it to converge to a single, rigid action pattern. Although performance remains acceptable on low-sparsity terrains, the policy cannot adapt to higher difficulty or greater randomization.

Fig. \ref{fig7} further highlights these differences at the highest difficulty level of stepping stones. The foothold distribution of PIE (START w/o TR-Net) shifts toward edges, exhibits greater variability and leads to frequent failures. Conversely, START proactively places footholds near stone centers similar as the upper-bound HMap GT, producing consistently concentrated distributions even under strong domain randomization, demonstrating precise, robust foothold selection.

The accuracy of terrain reconstruction is also essential for achieving high policy performance. Both TR-Net w/o Prop and TR-Net w/o GRU conduct explicit reconstruction but omit proprioceptive information or temporal memory. This prevents them from fully capturing terrain features and leads to inferior performance compared to START. Without U-Net refinement, although TR-Net w/o Refine can generate rough heightmaps that approximate safe foothold regions (e.g., stone and beam centers), it fails to delineate edges precisely. This inaccuracy increases MEV and reduces success rates on terrains like balance beams where edge precision is vital. Overall, START leverages high-fidelity local heightmaps to significantly enhance adaptability, robustness, and agility on challenging sparse footholds, even when relying solely on limited egocentric vision.

\begin{figure}[!tbp]
  \centering
  \vspace{-3pt}
  \includegraphics[width=0.96\linewidth]{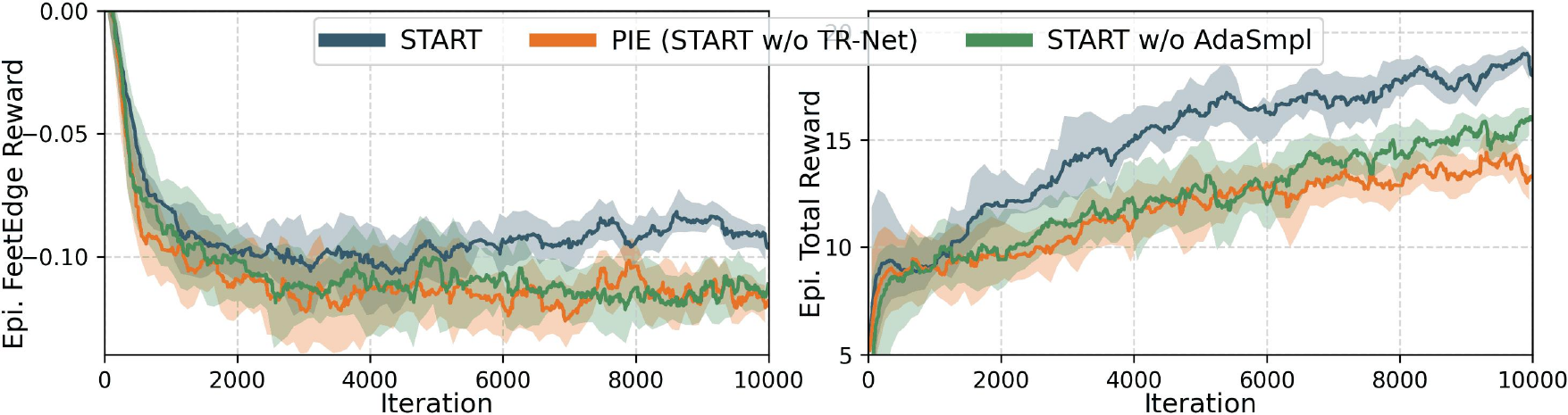}
  \vspace{-11pt}
  \caption{$r_{\text{feetedge}}$ and total reward in one episode for methods of START, PIE (START w/o TR-Net), and START w/o AdaSmpl. The former reflects the robot’s ability to select safe footholds. START achieves higher $r_{\text{feetedge}}$ and faster convergence, demonstrating the efficiency of the training framework.}
  \label{fig8}
  \vspace{-20pt}
\end{figure}

We also conducted ablation studies on different training pipelines. AdaSmpl plays a key role in improving exploration efficiency of locomotion policy. START w/o AdaSmpl is exposed to noisy, low-information reconstructions early in training, which hampers extraction and use of critical geometric cues like terrain edges. As illustrated in Fig. \ref{fig8}, the policy requires more training episodes to reach the same reward level as START, thereby degrading performance. Similarly, removing the foot-edge reward $r_{\text{feetedge}}$ (START w/o FtEdge) or the terrain progressive curriculum (START w/o TerProg) leads to significant performance drops. Lacking foot-edge guidance, START w/o FtEdge often places feet on edges, causing instability, falls, and extremely high MEV. Without progressive exposure from easier environments, START w/o TerProg trains directly on challenging terrains, where sparse rewards and conflicting optimal gaits hinder incremental skill discovery and result in overall failure. These results highlight another strength of START: a simple yet effective progressive learning pipeline that guides the policy from easy to hard tasks, enabling mastery of demanding locomotion skills and preserving superior adaptability across all risky terrains.

\begin{table*}[!htbp]
\setlength{\abovecaptionskip}{0pt}
\setlength{\belowcaptionskip}{0pt}
\caption{Ablation studies in the real world. Each method was tested 5 times with success rate and traversing rate recorded.}
\vspace{-7pt}
\label{table2}
\begin{center}
\renewcommand\arraystretch{1.22}
\begin{tabular}{llllllll}
\Xhline{0.75pt}
\noalign{\vskip 0pt}
\multicolumn{1}{c}{\multirow{2}{*}{Ablations}} &
  \multicolumn{2}{@{\hspace{-1pt}}c@{\hspace{-1pt}}}{\setlength{\fboxsep}{1pt}\colorbox{gray!15}{\strut\textbf{\emph{Stepping Stones}}}\ {\footnotesize{76.3\% sparsity}}} &
  \multicolumn{2}{@{\hspace{-1pt}}c@{\hspace{-1pt}}}{\setlength{\fboxsep}{1pt}\colorbox{gray!15}{\strut\textbf{\emph{Balance Beams}}}\ {\footnotesize{0.2m width}}} &
  \multicolumn{2}{@{\hspace{-1pt}}c@{\hspace{1pt}}}{\setlength{\fboxsep}{1pt}\colorbox{gray!15}{\strut\textbf{\emph{Stepping Beams}}}\ {\footnotesize{58.8\% sparsity}}} &
  \multicolumn{1}{@{\hspace{1pt}}c@{\hspace{2pt}}}{\setlength{\fboxsep}{1pt}\colorbox{gray!15}{\strut\textbf{\emph{Gaps}}}\ {\footnotesize{0.7m width}}} \\
\noalign{\vskip 0pt}
\cline{2-8}
\multicolumn{1}{c}{} &
Success Rate & Traversing Rate & Success Rate & Traversing Rate & Success Rate & Traversing Rate & Success Rate \\
\hline
\textbf{START} & \multicolumn{1}{c}{\textbf{1.0}} & \multicolumn{1}{c}{\textbf{1.00±0.00}} & \multicolumn{1}{c}{\textbf{0.8}} & \multicolumn{1}{c}{\textbf{0.99±0.03}} & \multicolumn{1}{c}{\textbf{1.0}} & \multicolumn{1}{c}{\textbf{1.00±0.00}} & \multicolumn{1}{c}{\textbf{1.0}} \\
\noalign{\vskip -1.5pt}
PIE \scriptsize{(START w/o TR-Net)} & \multicolumn{1}{c}{0.2} & \multicolumn{1}{c}{0.58±0.27} & \multicolumn{1}{c}{0.6} & \multicolumn{1}{c}{0.84±0.21} & \multicolumn{1}{c}{0.4} & \multicolumn{1}{c}{0.66±0.32} & \multicolumn{1}{c}{0.8} \\
\noalign{\vskip -1.5pt}
TR-Net w/o Prop & \multicolumn{1}{c}{0.4} & \multicolumn{1}{c}{0.89±0.19} & \multicolumn{1}{c}{0.4} & \multicolumn{1}{c}{0.75±0.19} & \multicolumn{1}{c}{0.6} & \multicolumn{1}{c}{0.85±0.25} & \multicolumn{1}{c}{0.6} \\
\noalign{\vskip -1.5pt}
TR-Net w/o GRU & \multicolumn{1}{c}{0.2} & \multicolumn{1}{c}{0.53±0.28} & \multicolumn{1}{c}{0.2} & \multicolumn{1}{c}{0.74±0.25} & \multicolumn{1}{c}{0.2} & \multicolumn{1}{c}{0.42±0.33} & \multicolumn{1}{c}{0.4} \\
\noalign{\vskip -1.5pt}
TR-Net w/o Refine & \multicolumn{1}{c}{0.4} & \multicolumn{1}{c}{0.74±0.27} & \multicolumn{1}{c}{0.6} & \multicolumn{1}{c}{0.92±0.15} & \multicolumn{1}{c}{0.4} & \multicolumn{1}{c}{0.71±0.34} & \multicolumn{1}{c}{0.2} \\
\noalign{\vskip -1.5pt}
START w/o AdaSmpl & \multicolumn{1}{c}{0.8} & \multicolumn{1}{c}{0.97±0.06} & \multicolumn{1}{c}{0.6} & \multicolumn{1}{c}{0.89±0.16} & \multicolumn{1}{c}{0.8} & \multicolumn{1}{c}{0.99±0.03} & \multicolumn{1}{c}{0.8} \\
\noalign{\vskip -1.5pt}
START w/o FtEdge & \multicolumn{1}{c}{0.6} & \multicolumn{1}{c}{0.75±0.34} & \multicolumn{1}{c}{0.4} & \multicolumn{1}{c}{0.78±0.21} & \multicolumn{1}{c}{0.4} & \multicolumn{1}{c}{0.78±0.31} & \multicolumn{1}{c}{0.0} \\
\noalign{\vskip -1.5pt}
START w/o TerProg & \multicolumn{1}{c}{0.2} & \multicolumn{1}{c}{0.55±0.37} & \multicolumn{1}{c}{0.4} & \multicolumn{1}{c}{0.87±0.14} & \multicolumn{1}{c}{0.6} & \multicolumn{1}{c}{0.80±0.28} & \multicolumn{1}{c}{0.4} \\
\Xhline{0.75pt}
\end{tabular}
\end{center}
\vspace{-19pt}
\end{table*}

\begin{figure*}[!htbp]
  \centering
  \includegraphics[width=0.96\linewidth]{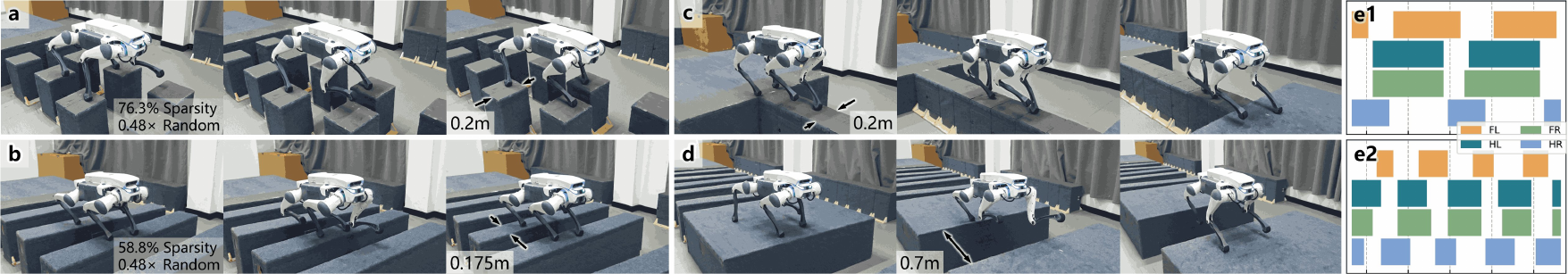}
  \vspace{-10.5pt}
  \caption{(a-d): Key frames of the robot traversing sparse foothold terrains in the real world, demonstrating agile, stable locomotion and sim-to-real generalization. (e): The robot exhibits different gait patterns on stepping stones and balance beams. (a)(e1): The robot traverses stepping stones slowly at low step frequency, placing each foot centrally within the sparse foothold area. (b)(e2): It first aligns its body over the beam center, then rapidly shifting the legs into a dynamically stable inverted-triangle stance and crossing the balance beam at high step frequency to counteract external disturbances.}
  \label{fig9}
  \vspace{-18pt}
\end{figure*}

\begin{figure}[!tbp]
  \centering
  \vspace{-3pt}
  \includegraphics[width=0.97\linewidth]{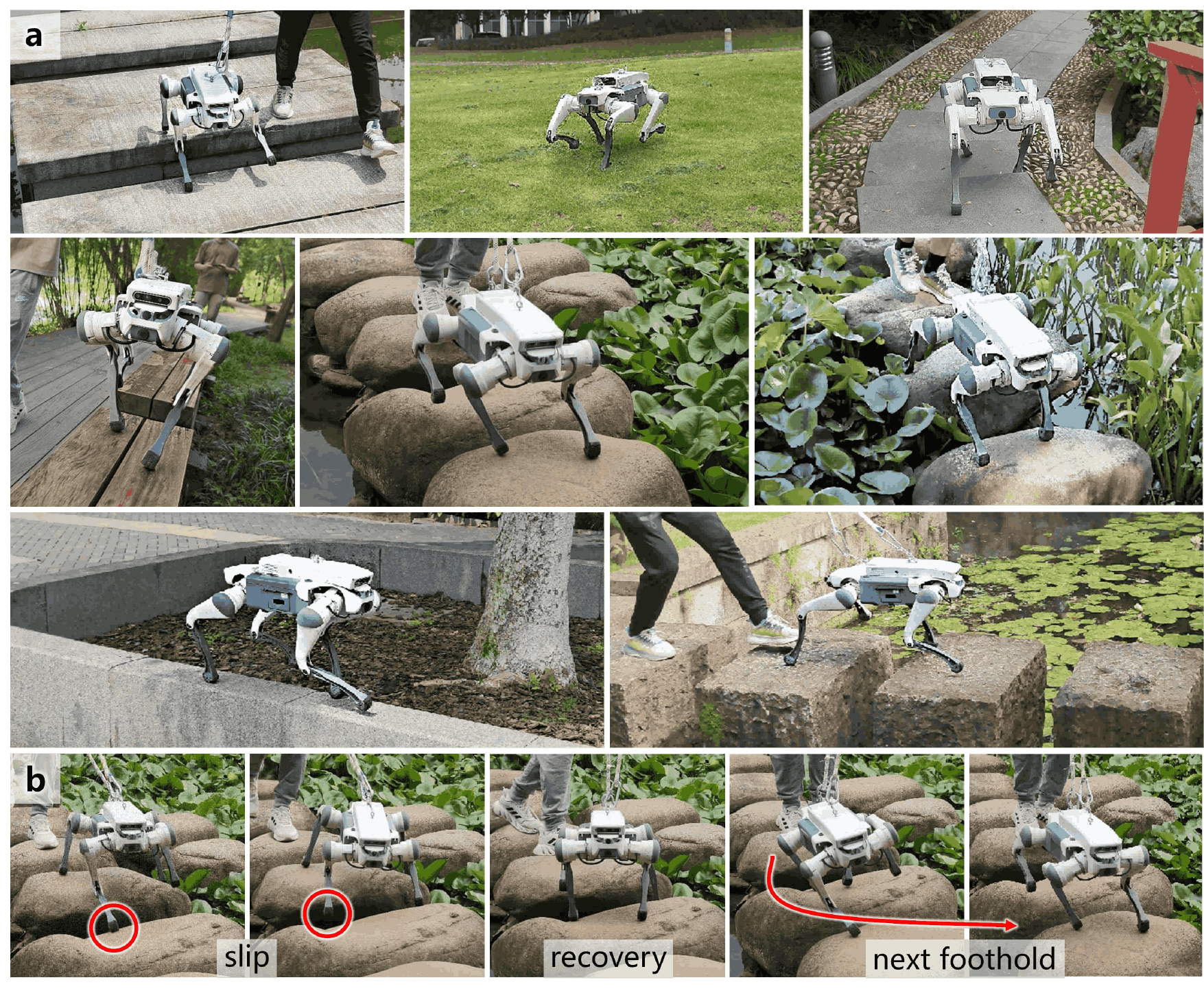}
  \vspace{-13pt}
  \caption{(a) We conduct outdoor experiments to validate START's locomotion performance in diverse complex sparse foothold terrains, including grassy ramps, park benches, irregularly shaped single-column stepping stones over water, and narrow (0.15m wide) slippery flowerbed edges, etc. (b) When confronted with uneven-surface stones never seen during training, the robot recovers rapidly from a slip and redirects to the next safe foothold.}
  \label{fig10}
  \vspace{-22pt}
\end{figure}

\vspace{-10pt}
\subsection{Real-World Lab-Level Experiments}

We use DEEP Robotics Lite3 quadruped robot for real-world deployment. An Intel RealSense D435i camera on its head captures raw depth at 10 Hz. We apply decimation, temporal, and hole-filling filters from the librealsense utilities to obtain $60\times 60$ images. Network inference runs on an onboard Jetson Orin NX with TensorRT acceleration, outputting action at 50 Hz. The learned policy can zero-shot transfer to complex real-world environments, as shown in Fig. \ref{fig9}.

To further validate the effectiveness of the proposed framework, we built four real-world terrains and compared START against all ablation methods (HMap GT excluded due to the lack of MoCap device), as summarized in Table \ref{table2}. Results show that PIE (START w/o TR-Net) suffers severe degradation in real world tests, experiencing missteps, edge placements, or directional deviations that lead to failures. Similarly, START w/o TerProg, trained directly on risky sparse foothold terrains, lacks adaptiveness and often attempts infeasible actions in the real world. In contrast, START leverages accurate explicit terrain reconstruction to match simulation performance across all terrains and robustly outperforms other methods. It traverses challenging sparse footholds at speeds up to 1.5m/s, continuously adjusting its gait to secure stable footholds. To our knowledge, no existing quadruped robot relying solely on onboard egocentric depth images has demonstrated comparable capabilities.

\vspace{-10pt}
\subsection{Real-World Outdoor Experiments}

We also conducted extensive experiments in outdoor natural sparse foothold scenarios, as shown in Fig. \ref{fig10}. Notably, although our method was never exposed to single-column stepping stones with uneven surfaces and irregular shapes, or vegetation obstructions during training, the policy's precise terrain perception enables safe foothold detection and stable foot placement, demonstrating excellent robustness and zero-shot generalization. Moreover, without specialized fine-tuning, START seamlessly adapts to rugged, non-flat terrains such as grassy ramps and descending staircases, highlighting its exceptional balance and adaptability.

\vspace{-8pt}
\section{CONCLUSION}

Rapid, accurate exteroceptive perception and dynamic gait adjustment are essential for traversing risky sparse foothold terrains. In this work, we present START, an end-to-end learning framework that, for the first time, enables agile quadruped locomotion with high traversability on highly sparse and randomized terrains using only low-cost onboard egocentric vision and proprioception. START employs local terrain heightmaps as an explicit intermediary to convey critical geometric features for stable locomotion. This representation allows the robot to infer rapidly varying surrondings and generate precise foothold placements, enhancing adaptability and stability. START also integrates the AdaSmpl method and the terrain progressive curriculum to unify all training components into a single-stage pipeline for improving exploration efficiency. Extensive simulation and real-world experiments demonstrate START’s robustness and generalization across diverse challenging terrains and its zero-shot sim-to-real transfer to complex real-world environments.

Although only four representative terrains were evaluated, START can extend to more generalized complex scenarios (e.g. stairs or boulders with height variations) and their combinations. In future work, we will extend START to more diverse settings to better cover real-world complexity and improve locomotion performance. To enhance environmental understanding, we will introduce finer edge-preservation mechanisms in heightmap refinement to better retain critical terrain features. We further plan to explore more expressive intermediate representations, such as voxels, that encode terrain geometry with higher fidelity and completeness.

\vspace{-6pt}

\bibliographystyle{IEEEtran}
\bibliography{IEEEabrv,ref}
\end{document}